\documentclass{article}
\usepackage{ijcai25}

\usepackage{times}
\usepackage{soul}
\usepackage{url}
\usepackage[hidelinks]{hyperref}
\usepackage[utf8]{inputenc}
\usepackage[small]{caption}
\usepackage{graphicx}
\usepackage{amsmath}
\usepackage{amsthm}
\usepackage{booktabs}
\usepackage{algorithm}
\usepackage{algorithmic}
\usepackage[switch]{lineno}
\usepackage{multirow}
\usepackage{booktabs}

\usepackage{amssymb}
\title{Correctness Learning: Deductive Verification Guided Learning for Human-AI Collaboration}

\author{
Zhao Jin$^1$
\and
Lu Jin$^1$\and
Yizhe Luo$^{1}$\and
Shuo Feng$^1$\and
Yucheng Shi$^1$\and
Kai Zheng$^2$\and
Xinde Yu\thanks{Corresponding authors}$^{1}$\And
Mingliang Xu$^{*1}$\\
\affiliations
$^1$Zhengzhou University\\
$^2$University of Electronic Science and Technology\\
\emails
\{jinzhao, luoyizhe, fengshuo, ieycshi, iexumingliang\}@zzu.edu.cn,
\{j18790929616, yxdzzu2022\}@163.com,
zhengkai@uestc.edu.cn
}

\begin{document}

\maketitle

\begin{abstract}
      Despite significant progress in AI and decision-making technologies in safety-critical fields, challenges remain in verifying the correctness of decision output schemes and verification-result driven design.
      We propose correctness learning (CL) to enhance human-AI collaboration integrating deductive verification methods and insights from historical high-quality schemes. The typical pattern hidden in historical high-quality schemes, such as change of task priorities in shared resources, provides critical guidance for intelligent agents in learning and decision-making.      
      By utilizing deductive verification methods, we proposed patten-driven correctness learning (PDCL), formally modeling and reasoning the adaptive behaviors—or ``correctness pattern''—of system agents based on historical high-quality schemes, capturing the logical relationships embedded within these schemes. Using this logical information as guidance, we establish a correctness judgment and feedback mechanism to steer the intelligent decision model toward the ``correctness pattern" reflected in historical high-quality schemes.
      Extensive experiments across multiple working conditions and core parameters validate the framework's components and demonstrate its effectiveness in improving decision-making and resource optimization.
\end{abstract}

\section{Introduction}

With the growing adoption of intelligent decision-making support systems (IDSS) in safety-critical domains such as smart manufacturing 
 (\cite{li2022review}), transportation (\cite{visan2022towards}), and electricity management (\cite{mansouri2023iot}), the demand for enhanced system correctness, reliability and trustworthiness has increased. 
 
  The correctness of IDSS generally refers to its ability to produce accurate outputs for valid inputs. ``Correctness'' is defined relative to human judgment (\cite{sokol2024does}), assuming that humans can make correct judgments. Thus, an IDSS is considered correct if and only if, for any input, the judgments made by humans and the system are identical. 
 Reinforcement learning from human feedback (RLHF) guides the agents to converge in the right direction by providing interactive feedback to the decision model\cite{kaufmann2023survey}.
However, human evaluation on decision-making outcomes in complex logical tasks is often imprecise, limited by individual experience, preferences, and intuition.
These limitations pose significant challenges for effectively training decision models.

\begin{figure*}[tbh]
		\centering
		\includegraphics[width=18cm,height=9cm]{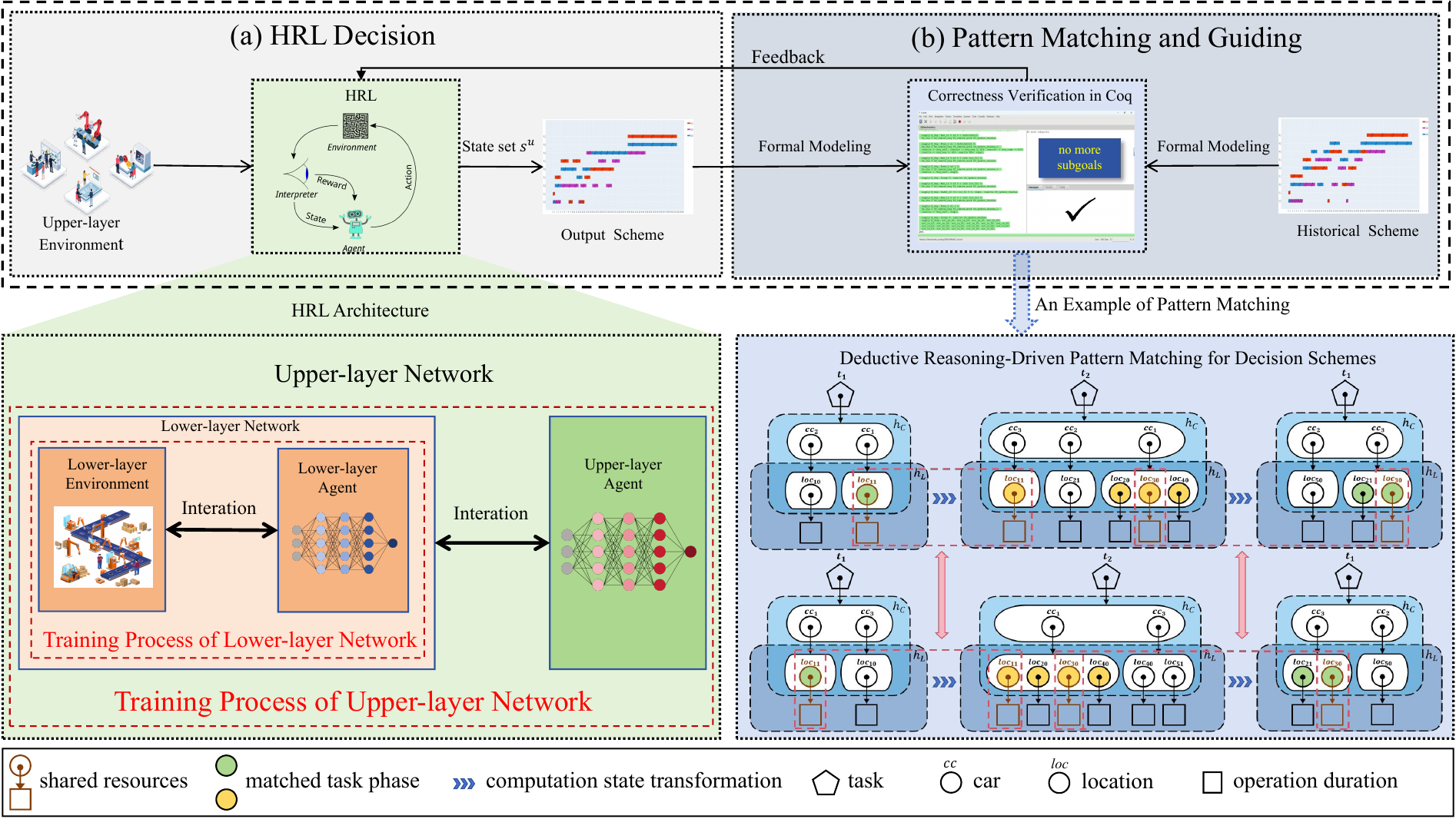}
		\caption{ Overview of the proposed correctness learning framework PDCL. In the first stage, a hierarchical reinforcement learning (HRL) model is constructed to generate output schemes. In the second stage, typical correctness patterns (task priorities) of historical high-quality schemes and output schemes are formally proved in Coq and feedback is given to the model based on the degree of pattern matching.
}\label{1}
	\end{figure*}

Formal verification constitutes a rigorous mathematics-based technology for the specification, verification, and design of computation systems, serving as a crucial approach to enhance and ensure the correctness of these systems.
Formal verification techniques are primarily categorized, according to methodological classifications, into deductive verification, model checking, and abstract interpretation.
Integrating formal verification into IDSS has emerged as a significant research focus (\cite{verma2019imitation,beard2022black,hunt2021verifiably,kouvaros2023towards,ghosh2023interpretability}).
 Recent efforts have explored verifying intelligent decision-making algorithms using formal verification techniques such as model checking (\cite{krichen2022formal,fu2018model}) and abstract interpretation (\cite{landers2023deep,albarghouthi2021introduction}). 
However, these studies have yet to fully address the verifiability challenges posed by IDSS in decision-making contexts, including the need for more complete correctness proofs of decision output schemes, the implementation of compositional reasoning for intelligent decision-making, and the development of a formal-specification-driven design framework.
 
 As another formal verification technique, deductive verification employs formal languages, semantics, logical reasoning, and theorem proving tool to ensure the correctness of computation systems. Beyond verification, this technique facilitates a deep understanding of the system behavior\cite{wolfman1999lpsat}, which in turn aids in guiding the design of the system.
 Adapting and applying the deductive reasoning techniques to the field of intelligent decision-making is an alternative way to deal with the problems mentioned above.
 However, formal verification approaches based on deductive reasoning remain relatively underexplored in IDSS. This limitation arises because deductive verification requires a formal definition of correctness and a clear understanding of the principles underlying the target system's operation, both of which remain unresolved in this field.

 To bridge this gap, we focus on ensuring the correctness of output instructions generated by IDSS while providing precise descriptions and specifications of the behavior of agents, rather than formalizing the inherently complex decision-making process. 
 By doing so, we introduce deductive verification into intelligent decision-making models for the first time, and propose correctness learning (CL) with the following core concept: verifying the correctness of the decision output instructions (i.e., decision schemes). Simultaneously, it deduces, specifies, and symbolically describes the hidden and cumulative behavior of agents from the execution process of the instructions, and then sets a benchmark with respect to the formalized agents' behavior to guide the evolution of the learning model along with the expected direction.




 

In sight of the above, we propose a pattern-driven correctness learning (PDCL): taking historical high-quality schemes as the benchmark for correctness, by identifying the behavioral patterns in historical high-quality schemes, the "correctness pattern" in historical schemes is injected into the decision-making model to improve performance.
Specifically, first, we decouple the process decision from resource allocation, executing them separately at the upper and the lower layers. At the lower layer, we train a resource allocation network, which, upon convergence, is integrated into the upper-layer environment to provide real-time feedback for the process decision network. Simultaneously, the process decision network is trained to handle scheduling processes, ultimately generating complete scheduling schemes.
 Second, we conduct deductive verification of historical high-quality schemes utilizing the mathematical reasoning capabilities of separation logic for shared resource allocation and management. By modeling, analyzing, and reasoning the scheme execution, we symbolically describe the typical ``pattern" such as local resource utilization, allocation and release, resource occupation, and task prioritization.
 Third, deductive verification ensures the correctness of decision output schemes of the decision model and evaluates the extent to which the intermediate output schemes of the decision model exhibiting the typical pattern match historical high-quality schemes. A reward mechanism, based on the principle that a higher matching degree with these patterns yields higher rewards, incentivizes the model to adjust toward the desired direction. This approach enables the model to surpass traditional scheduling methods, achieving more efficient resource allocation.

 Overall, the contributions of this paper are as follows:
 \begin{itemize}
\item We are the first to introduce deductive verification into IDSS, and propose correctness learning, accomplishing the verification of decision schemes and the symbolic description of the behavior of agents, by which 
establishing a novel learning mechanism.
 
\item We propose a pattern-driven correctness learning, which effectively combines the insights derived from historical high-quality experiences with the robust exploratory capabilities of intelligent learning algorithms. 

\item  We extensively evaluate PDCL across different working conditions, the performance of the four benchmark algorithms improved by an average of 8.4\%, 3.9\%, 1.6\%, and 5.7\%, respectively, and we also analyze the impact of core parameters on the effectiveness of PDCL.

\end{itemize}

\section{Related Work}

\subsection{Reinforcement Learning for Human Feedback}
\cite{christiano2017deep} introduced human experience to assist model learning in deep reinforcement learning. Subsequently, several scholars studied the intricate interaction between human experience and intelligent agents\cite{kaufmann2023survey,chakraborty2024parl,zhang2024multi,crochepierre2022interactive}. We focus on the sources of feedback information for intelligent agents, including large language models (LLMs), human experience, and logical reasoning. \cite{brooks2023large} implemented policy iteration using a contextual learning mechanism over an LLM, enabling it to perform reinforcement learning tasks without expert demonstrations or gradients. \cite{cao2024enhancing} leveraged logical rules derived from the environment to decompose task goals and guide agents to enhance human-AI perception and collaboration. Compared to the above methods, we alleviate the inherent limitations of current machine learning methods in terms of explainability by embedding verification based on deductive reasoning into the learning process.

\subsection{Safe Reinforcement Learning}
Safety reinforcement learning is divided into constrained reinforcement learning\cite{bai2023achieving,bharadhwaj2020conservative}, shielded reinforcement learning\cite{carr2023safe,beard2022black,odriozola2023shielded} and risk-constrained policy gradients \cite{xiao2024policy}. Shielded reinforcement learning combines formal verification with reinforcement learning and introduces ``shielding" to verify whether the agent's behavior is safe and check shield dangerous behaviors. The above work only focused on process reliability, limiting the decision path to a range of satisfying properties, and did not consider using theorem-proving methods to embed the correctness of decision results to help improve decision-making effects.

\section{Problem Formulation}
In this section, we formally define scenario modeling and introduce it in detail. 

\noindent\textbf{Definition 1} (Operation). An operation is a two-entry tuple $o=(w, t)$, where $w$ represents the resources required by the operation and $t$ represents the execution duration of the operation. 

\noindent\textbf{Definition 2} (Task). A task is a two-entry tuple $\tau=(O,\beta)$, where $O$ is a set representing the operations required to complete a task, and the set of operations contained in a task must be executed sequentially. $\beta$ represents the progress of task completion.

\noindent\textbf{Definition 3} (Equipment). An equipment can be represented as a two-entry tuple $e=(w, n)$, where $w$ represents the type of resources that the equipment can provide and $n$ represents the number of workstations that the equipment can provide.

\noindent\textbf{Definition 4} (Car). A car is a two-entry tuple $c=(l,b)$, where $l$ represents the location and $b$ represents whether it is available.

\noindent\textbf{Definition 5} (Job ). Given a set of tasks $\Gamma$ associated with a set of $O$, a set of $C$ and a set of equipment $E$, job scheduling assigns appropriate devices $e\in E$ to perform a job based on the resources required by operation $o\in O$ of task $\tau \in \Gamma$. Its goal, $T^g$ is to minimize the total completion time of each batch of tasks, which can be defined as
\begin{equation}\begin{aligned}
T^g=\min\sum_{\tau\in\Gamma}\sum_{o\in O}\sum_{e\in E}T^e(\tau,o)+T^w(\tau,e\lor c)
\end{aligned}\end{equation}
where $T^e$ represents the time required for the task to execute the job, and $T^w$ represents the time the task waits for the allocation of equipment or car.

\section{Algorithm}
In this section, we describe the construction of hierarchical reinforcement learning, deductive verification, and model training in detail. The overview is shown in Figure\ref{1}.

\subsection{Hierarchical Reinforcement Learning}

\subsubsection{Lower-layer Model}
\textbf{States} $s^l$: The state space of the lower-layer model at time $t$ is defined as: $s^l=(\tau_t, E^l_t, C^l_t)$, where $\tau_t$ represents the state matrix of one task, $E^l_t$ represents the state matrix of the equipments, $C^l_t$ is the state matrix of the cars. The initial state of the car is randomly generated.

\noindent\textbf{Actions} $a^l$: The lower-layer model selects the most suitable car from multiple available cars to complete the task. The action space is defined as $a^l_t\in\{0,1,2,\cdots,i,\cdots,K\}$. The action $a^l_t$ = 0 indicates to keep waiting and not select a car, $a^l=i$ indicates the specific serial number of selected cars, and $K$ is the total number of cars.

\noindent\textbf{Reward function} $r^l$: The lower-layer model reward function consists of the task reward $r^l_x$ and the process reward $r^l_y$, expressed as $r^l=r_p+r_q$, where $r_p$ is the task reward $\mathcal{R}$ obtained when the task reaches the target location, and $\mathcal{R}$ is a constant. $r_q$ is the immediate feedback during the task and is defined as
\begin{equation}r_q=
\begin{cases}
2, & \exists C^l_{d,t}=1\land a^l_t=d \\
1, & \forall C^l_t=-1\land a^l_t=0 \\
-2, & \exists C^l_t=1\land a^l_t=0 \\
-2, & \forall C^l_t=-1\land a^l_t\neq0 \\
-2, & \exists C^l_t=1\land C^l_{d,t}=-1\land a^l_t=d
\end{cases}\end{equation}
1) If both the equipment and car are idle and an action is assigned to the agent, a reward is given. 2) If there is no idle time for the equipment or car when the agent chooses to wait, a rewarded is given. 3) If the agent chooses to allocate when there is no idle time on the equipment or car, a penalty is imposed. 4) If there is no idle time for equipment and cars, but actions are assigned to the agent, a penalty is imposed. 5) If the cars are idle, a non-idle car is assigned to the agent, and a penalty is imposed.

\subsubsection{Upper-layer Model}
\textbf{States} $s^u$: The state space of the upper-layer model at time $t$ is defined as: $s^u=(\Gamma_t, E^u_t, C^u_t, F_t)$, where $\Gamma_t$ represents the state matrix of tasks, $E^u_t$ represents the state matrix of the equipment, $C^u_t$ is the state matrix of the car, and $F_t$ is used to describe the completion progress of job scheduling.

\noindent\textbf{Actions} $a^u$: The action space contains the set of task scheduling decisions that can be executed in different states. It is defined as $a^u \in \{1,\cdots,i,\cdots,N\}$. Among them, $a^u_t= i$ means that the i-th task selected in a given state is assigned to the equipment.

\noindent\textbf{Reward function} $r^u$: $r^u$: The upper-layer model focuses on the overall task completion, and the reward value is defined as $r^u_t=r_x+r_y+r_z$. $r_x$ is determined by the maximum time spent on each decision-making process. The smaller the time difference between two consecutive decisions, the higher the parallelism of the two tasks on the timeline. This reward mechanism encourages the agent to increase task parallelism to improve overall task efficiency. $r_x$ is defined as:
\begin{equation}r_x=\frac{M^{old}-M^{new}}{5000}\end{equation}
$M$ represents the maximum value of the overall time step of the system after the execution of the job scheduling mission. This parameter reflects the maximum time required at the beginning of the mission. 

$r_y$ is used to constrain the decision-making behavior and impose penalties when the task reaches the target location and is still selected. It is defined as:
\begin{equation}\begin{aligned}
r_y=-1-\frac{(\mathcal{N}-F_t)}{\mathcal{N}}
\end{aligned}\end{equation}
$\mathcal{N}$ is a constant used to represent the completion progress of job scheduling. 

After each round of training, if the scheme obtained can complete all tasks, we will verify the current scheme. By comparing the output scheme with the historical high-quality scheme, get $r_z$ according to the matching degree, defined as:
\begin{equation}\begin{aligned}
r_z=-\alpha*(\mu_{match}-\mu_{total}/2)
\end{aligned}\end{equation}
where $\mu_{match}$ represents the matching degree between the pattern obtained in the output scheme and the historical high-quality scheme, and $\alpha$ is the weight coefficient.

\begin{figure}[!t]\centering
		\includegraphics[width=8.5cm,height=3cm]{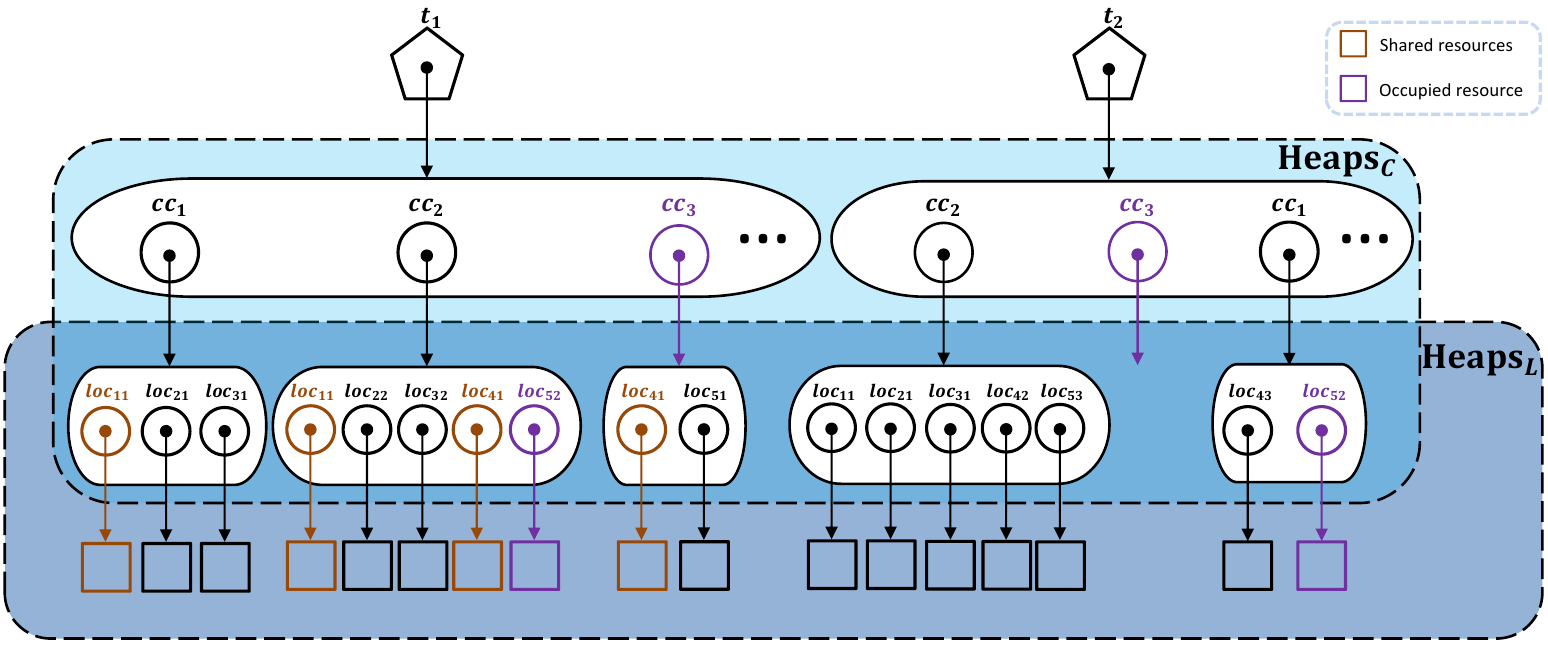}
		\caption{Schematic diagram of the two-tier resource heap model of the job scheduling system.}\label{2}
	\end{figure}

\subsection{Deductive Verification}

\subsubsection{Formal model of the job scheduling system}
Based on separation logic, the behavior of the job scheduling system is formally modeled as a two-layer, resource-separated, two-tier heap structure model, as shown in Figure\ref{2}.
The core components of this model are the ``car resource heap" (i.e., $\mathrm{Heaps}_C$ in the Figure\ref{2}) and the ``location resource heap" (i.e., $\mathrm{Heaps}_L$ in the Figure\ref{2}, the location of the equipment). The car resource heap establishes a relationship between the ``mapping from task to car" and the ``mapping from the car to location" while ensuring the mutual independence of the car resources. Conversely, the location resource heap is used to describe the underlying operation mode of the transport task, in which each location represents a shared deployable resource. By doing so, this model possesses sufficient expressive power to describe the behavior of the job scheduling system, encompassing 1) the interrelationship between the two types of resources, 2) the sequence of workflow execution process established within the scheme, 3) cooperative scheduling among multiple tasks, and 4) the prolongation of resource occupancy resulting from extensive resource sharing.
\begin{figure}[tbh]
		\centering
		\includegraphics[width=8.5cm]{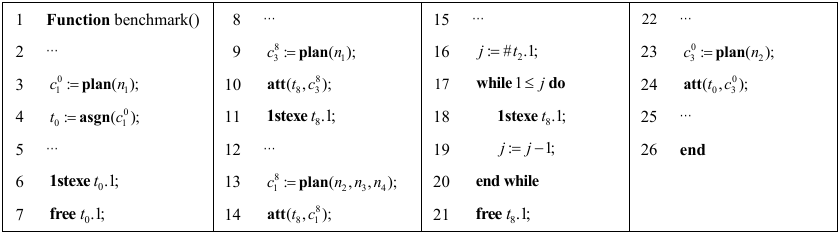}
		\caption{Modeling program of the historical high-quality schemes (simplified)}\label{8}
	\end{figure}
\subsubsection{Modeling language of the job scheduling system}
Based on the aforementioned two-tier resource heap model, a modeling language for the job scheduling system called $\mathbf{ML}_{\mathbf{JSS}}$ is constructed. It primarily describes the resource allocation and management operations during the actual execution of the operation scheme. In the following, the syntax of the modeling language is defined, along with its computational state and operational semantics.

\noindent\textbf{Definition 6}. The full syntax of the expressions and commands of $\mathbf{ML}_{\mathbf{JSS}}$ is given as follows:

\begin{equation}
\begin{aligned}
&e:=n,m,...\mid x,y,...\!\mid\! e_{1}+e_{2}\!\mid\! e_{1}-e_{2}\!\mid\! e_{1}\times e_{2}\!\mid\!\#ce\!\mid\!\#t\\
&be:=e_{1}\!=\!e_{2}\!\mid\! e_{1}\!\leq \!e_{2}\!\mid\!\mathbf{true}\mid\!\mathbf{false}\mid\!\neg be\mid\!  be_{1}\!\vee\! be_{2}\!\mid\!  be_{1}\!\wedge\! be_{2}\\
&te:=\mathbf{null}\!\mid\!\mathbf{fin}\!\mid \!t_{1},t_{2},...\!\mid \!te\cdot ce\!\mid \!te_{1}\cdot te_{2}\\
&ce:=\mathbf{null}\mid n,m,...\mid c_1^a,c_2^a,...\mid t.e\\
&C:=x:=e \!\mid \!
C;C^{\prime} \!\mid \! \mathbf{if}\ be\ \mathbf{then}\ C\ \mathbf{else}\ C^{\prime} \!\mid \! \mathbf{while}\ be\ \mathbf{do} \ C^{\prime}\\
&\ \ \ \ \ \ \ \ \ |t:=\mathbf{asgn}\left(ce^{*}\right)|\mathbf{att}\left(t,ce^{*}\right)|\mathbf{free}\ t.e|\mathbf{comp}\ t\\
&\ \ \  \ \ \ \ \ \  |c^a:=\mathbf{plan}\left(\overline{e}\right)\ \!|\!\ \mathbf{add}\left(c^a,e\right)\ \!|\!\ x:=\{ce.e\}|\mathbf{1stexe}\ t.e\ 
\end{aligned}
\end{equation}
where $e$ is written for location expressions, $be$ for Boolean expressions, $te$ for task expressions, $ce$ for car expressions, and $C$ for commands (including the standard IMP command inherited in the first line and the newly added task and car operation commands in the last two lines).

The computational state of the modeling language $\mathbf{ML}_{\mathbf{JSS}}$ is defined as a quintuple, i.e., a state $\sigma\in\mathrm{States}$ is of the form $(s_{T},s_{C},s_{L},h_{C},h_{L})\in\mathrm{Stores}_{T}\times\mathrm{Stores}_{C}\times\mathrm{Stores}_{L}\times\mathrm{Heaps}_{C}\times\mathrm{Heaps}_{L}$.

\noindent\textbf{Definition 7}. The operational semantics for the plan command of $\mathbf{ML}_{\mathbf{JSS}}$ is given below as an example. The complete semantics is given in Appendix B.

\begin{equation}
		\begin{tabular}{c}
			$cc\in\mathcal{C}-\mathrm{dom}(h_{C})\text{and}loc_{1},...,loc_{n}\in\mathrm{Loc}-\mathrm{dom}(h_{L})$\\
			\hline
			$\langle c^a:=\mathbf{plan}\left(\overline{e}\right),\sigma\rangle\leadsto(s_{T},[s_{C}\mid c^a:cc],s_{L},[h_{C}\mid cc:$\\$(loc_{1},...,loc_{n})],[h_{L}\mid loc_{1}:[[e_{1}]]\sigma,...,loc_{n}:[[e_{n}]]\sigma])$ 
		\end{tabular}
	\end{equation}
Intuitively, this command plans the process undertaken by car $c$; that is, it assigns a location sequence to the car based on the order of the workflow, and the corresponding operation duration value of each location constitutes sequence $\bar{e}$.

\begin{table*}[ht]
	\centering
	\caption{Test results of different algorithms and different algorithms with rule-guided strategies across ten task scenarios and twelve task scenarios. Four indicators are used for evaluation: completion time (ComT, lower the better), cumulative reward (CumR, higher the better), decision time (DecT, higher the better), and training time (TraT).\label{tab:table1}}
	\setlength{\tabcolsep}{5mm}{
    \setlength{\tabcolsep}{10pt}
    \renewcommand\arraystretch{1.2}
		\begin{tabular}{ccccccccc}		
			\hline
           \multirow{2}{*}{\textbf{Method}} & \multicolumn{4}{c}{\textbf{Ten Tasks}} &   \multicolumn{4}{c}{\textbf{Twelve Tasks}}\\
            \cmidrule(r){2-5}\cmidrule(r){6-9}
			 &ComT $\downarrow$ & CumR$\uparrow$ & DecT$\downarrow$ &TraT &ComT $\downarrow$ & CumR$\uparrow$ & DecT$\downarrow$ &TraT\\
			\hline
			DQN &   64 & -0.0128 & 52.37 &430 & 68&  -0.0136  & 59.97 &524 \\
			
			DDQN &  60  & -0.0120 & 20.41 & 355 & 71& -0.0142 & 46.05 &1026 \\
			
			Dueling DQN & 58 &-0.0116 & 22.47  & 354& 68 &-0.0136 & 24.39 & 313\\
			
			PPO &  63 & -0.0126 & 79.72 &324 &80 & -0.0160  & 80.97 &417 \\
			
			\hline
			DQN \& PDCL &58 & -0.0116  &53.46  & 496&  63& -0.0126 & 64.24 & 930\\
            
                DDQN \& PDCL &57  & -0.0114 & 21.70 & 359& 69 & -0.0138 &  56.92& 1204 \\

                Dueling DQN \& PDCL&  57& -0.0114 &24.12 & 371 & 67& -0.0134 & 25.71 & 412 \\

                PPO \& PDCL & 59 &-0.0118  & 80.91 & 425& 76 &  -0.0152 & 83.10 &561 \\
			\hline
	\end{tabular}}
\end{table*}

\begin{table}[ht]
	\centering
	\caption{The test results of different algorithms and different algorithms with rule-guided strategies across ten task scenarios after adjusting the learning rate.\label{tab:table2}}
	{
    \renewcommand\arraystretch{1.2}
    \resizebox{0.45\textwidth}{!}{
		\begin{tabular}{ccccc}		
			\hline
			 \textbf{Method}&ComT $\downarrow$ & CumR$\uparrow$ & DecT$\uparrow$ &TraT  \\
			\hline
			DQN & 57 &-0.0114 & 67.13  &372 \\
			
			DDQN & 56   &  -0.0112& 40.07 & 890 \\
			
			Dueling DQN & 56 &-0.0112 & 66.17  &438 \\
			
			PPO & 75  & -0.0150 & 75.33 & 607 \\
			
			\hline
			DQN \& PDCL &56 &-0.0112 & 69.24  &382\\
            
                DDQN \& PDCL &56  & -0.0112  & 45.05 &1081 \\

                Dueling DQN \& PDCL & 56 &-0.0112 & 67.64  &971  \\

                PPO \& PDCL &74  & -0.0148 & 77.77 & 649\\
			\hline
	\end{tabular}}
    }
\end{table}

\subsubsection{Correctness verification of the schemes}
The correctness of a scheme can be ensured by describing the scheme as a modeling program by $\mathbf{ML}_{\mathbf{JSS}}$ and checking whether the execution behavior of this program is as expected according to the operational semantics of $\mathbf{ML}_{\mathbf{JSS}}$.
We subsequently demonstrate our verification method by proving the sample modeling program in Figure\ref{8}, which describes the dynamic implementation process of the historical high-quality schemes. The sample program involves the allocation and management of two types of resources, cars, and locations, also known as the resource consumption reference in the real-world engineering field.
To enhance readability, we isolate two of the shared resource sub-tasks in the scheme for demonstration. This is feasible because of the local reasoning strategy in separation logic, permitting our reasoning locally to focus only on the sub-resource heap that is mutated and ignore all others.
\begin{algorithm}[tb]
    \caption{Hierarchical reinforcement learning decision model training}
    \label{alg:algorithm}
    \textbf{Input}: The maximum number of training steps $T$, trained lower-layer policy network $\pi_{\theta_l}$, upper-layer policy network $\pi_{\theta_u}$ reward coefficient $\alpha$, query function $fun$, verification tools Coq, the historical high-quality scheme $h$,constant $\mathcal{N}$.\\
    \textbf{Output}: tuple ($s^u_t$, $a^u_t$, $r^u_{t}$, $s^u_{t+1}$).
    
    \begin{algorithmic}[1] 
        \WHILE{$T$}
        \STATE The upper-layer policy network $\pi_{\theta_u}$ selects an action $a^u_t$ based on the state $s^u_t$.
        \STATE Obtain the lower-layer state $s^l_t$ based on the upper-level action $a^u_t$ and the current state $s^u_t$.
        \STATE Use the trained lower-layer policy network $\pi_{\theta_l}$ to directly generate the action $a^l_t$.
        \STATE Execute the lower-layer action $a^u_t$ and obtain the next state $s^l_{t+1}$ from the lower-layer environment.
        \STATE Update the upper-layer state to get $s^u_{t+1}$ based on the lower-layer state $s^l_{t+1}$.
        \STATE Get rewards $r^u_t=r_x+r_y$
        \IF {$F_t=\mathcal{N}$}
        \STATE  $r^u_t=\alpha*fun(Coq(s^u_t),Coq(h))+r^u_t$
        \ENDIF
        \ENDWHILE
        \STATE \textbf{return} experience tuple ($s^u_t$, $a^u_t$, $r^u_t$, $s^u_{t+1}$)
    \end{algorithmic}
\end{algorithm}

\noindent \textbf{\textit{Proof sketch.}} Starting from a given initial state $(s_{T},s_{C},s_{L},$\\$h_{C},h_{L})$(Line 2), according to the order of the workflow in the scheme, a car $c_1^{0}$ is planned to perform an operation of the first phase of one task at an implicit location $loc_{11}$ with operation duration $n_1$(Line 3), and a task $t_0$ is subsequently assigned to this car (Line 4). By the operational semantics of the $\mathbf{plan}$ and $\mathbf{asgn}$ commands, the initial state is transitioned to 
$([s_{T}|t_0\!:\!cc_1],[s_{C}|c_1^{0}\!:\!cc_1],s_{L},[h_{C}|cc_1\!:\!(loc_{11})],[h_{L}|loc_{11}\!:$\\$\!n_1])$. 
We then execute the first item of the sequence of operations associated with task $t_0$, that is, removing the first item of the location sequence corresponding to the car resource $cc_1$ in $h_C$ and releasing the corresponding location resource $loc_{11}$ from $h_L$(Line 6). 
Subsequently, we release the car resource $cc_1$, whose content value is required to be an empty sequence $\mathbf{null}$, corresponding to the first operation of task $t_0$ from $h_C$, that is, $c_1^{0}$. The state is now $([s_{T}|t_0:\mathbf{null}],[s_{C}|c_1^{0}\!:\!cc_1],s_{L},h_{C},h_{L})$. 
In Line 9, a new car $c_3^{8}$ is planned for an operation of the first phase of one task at the same location $loc_{11}$ of the first operation of task $t_0$, which is previously released in the previous step, and then attach this car to the end of an existing task $t_8$ (Line 10), thus obtaining a state $([s_{T}|t_8:cc_3],[s_{C}|c_3^{8}\!:\!cc_3],s_{L},[h_{C}|cc_3\!:\!(loc_{11})],[h_{L}|loc_{11}\!:\!n_1])$. 
The first operation of task $t_8$ is then executed in Line 11 in a similar manner to Line 6, resulting in a state $([s_{T}|t_8\!:\!cc_3],[s_{C}|c_3^{8}\!:\!cc_3],s_{L},[h_{C}|cc_3\!:\!\mathbf{null}],h_{L})$. Subsequently, in Line 13, the released car $cc_1$ is activated again and is planned to undertake three consecutive phases of the task $t_8$, allocating three new location resource $loc_{20}$, $loc_{30}$ and $loc_{40}$, and their corresponding operation duration form a sequence $(n1,n2,n3)$. After attaching this car to task $t_8$ in Line 14, state $([s_{T}|t_8\!:\!cc_1],[s_{C}|c_1^{8}\!:\!cc_1],s_{L},[h_{C}|cc_1\!:\!(loc_{20},loc_{30},loc_{40})],[h_{L}|loc_{40}\!:\!n_4|loc_{30}\!:\!n_3|loc_{20}\!:\!n_2])$ is obtained.
Through the command between Line 16 and Line 20, involving a while loop, the three phases of the task $t_8$ are executed successively. After removing the corresponding car resource $c_1^{8}$ from task $t_8$ in Line 21, state $([s_{T}|t_8\!:\!\mathbf{null}],[s_{C}|$\\$c_1^{8}\!:\!cc_3],s_{L},h_{C},h_{L})$ is obtained. Subsequently, the second phase of the task $t_0$ is deployed in Line 23-24, resulting in a state $([s_{T}|t_0\!:\!cc_3],[s_{C}|c_3^{0}\!:\!cc_3],s_{L},[h_{C}|cc_3\!:\!(loc_{20})],[h_{L}|$\\$loc_{20}\!:\!n_2])$. The remainder describes the subsequent implementation of the scheme; therefore, it is not repeated.

In contrast to task $t_8$, which can be assigned and executed continuously in Lines 9-21, task $t_0$ was divided into two task phases. Thus, the first and second stages start at Lines 3 and 23, respectively, to meet the established process timing of the scheme. The transformation relation between configurations shows that the two task-running processes share two locations, namely, $loc_{11}$ and $loc_{20}$, and meet the established requirements of the scheme. Thus, the $t_0$ task at $loc_{11}$ has a higher sequential priority, whereas the $t_8$ task at $loc_{20}$ has a higher sequential priority. In the above verification process, we identified and symbolically represented the hidden phenomenon of the stable operation of the job scheduling system, such as the law of change in the priority of workflow under resource preemption.

\subsubsection{Formal Development in Coq}
Based on the above formalization, we implemented the modeling language $\mathbf{ML}_{\mathbf{ASS}}$ in proof assistant Coq, in which we interactively verified the correctness of the historical high-quality schemes(as described in Figure\ref{8}) and the intelligent decision model output scheme, and symbolically analyzed the pattern contained in the scheme. The development is available online at the following URL. A diagram of the Coq deployment is given in Appendix C.

With this tool, one can write the scheme to be verified into a modeling program and interactively prove the termination and execution correctness of this modeling program to ensure the correctness of the scheme. Moreover, during verification, the transformation relationship between the configurations of the execution process of operation commands can be analyzed based on the operational semantics of the corresponding commands. This symbolically describes the hidden and cumulative adaptive behavior properties and pattern of the subject within the operating system.

\subsection{Model Training}
As shown in Algorithm ~\ref{alg:algorithm}, the upper-layer model is trained to ensure the processing of a single task. At this stage, the low-layer model focuses on the car's allocation. Subsequently, the upper-layer module obtains state $s^l_t$ based on state $s^u_t$ and action $s^u_{t+1}$. The action is generated using the trained low-layer model; the bottom-level action is executed; the next state $s^l_{t+1}$ is obtained from the execution model environment; and the top-level state is updated with state $s^l_{t+1}$ to obtain $S^l_{t+1}$ and the reward value $r^u_t$. 
\begin{table}[ht]
	\centering
	\caption{The test results of different algorithms and different algorithms with rule-guided strategies across ten task scenarios after adjusting the activation function.\label{tab:table3}}
	{
    \renewcommand\arraystretch{1.2}
    \resizebox{0.45\textwidth}{!}{
		\begin{tabular}{ccccc}		
			\hline
			 \textbf{Method}&ComT $\downarrow$ & CumR$\uparrow$ & DecT$\uparrow$ &TraT  \\
			\hline
			DQN & 57  & -0.0114&50.98  & 799 \\
			
			DDQN & -   & -1.0453 & 50.46 & 475 \\
			
			Dueling DQN & -   & -1.1588 & 52.94 & 423\\
			
			PPO& 61  & -0.0122 & 78.93 & 226  \\
			
			\hline
			DQN \& PDCL & 56  & -0.0112&59.82&856 \\
            
                DDQN \& PDCL & -   & -1.0616 & 53.49 & 627\\

                Dueling DQN \& PDCL&  56&-0.0112  &66.57 & 648  \\

                PPO \& PDCL & 59 & -0.0118 & 80.75 &227\\
			\hline
	\end{tabular}}
    }
\end{table}
To better conduct interactive training, we abstract the resource priority transformation phenomenon found in the deductive reasoning process into rules to verify the degree of pattern matching between the output plan and the historical high-quality plan. The design of the reward mechanism follows the principle that the more rules are matched, the greater the reward value, so that the model breaks through the original scheduling plan and achieves more efficient resource allocation.

\section{Experimental Evaluation}
\subsection{Experimental Setting}

\noindent\textbf{Simulation environment.} Simulation environment reference \cite{yujie2023optimization} construction. In the simulation scenario, each task consists of five operations that must be executed in sequence. Equipment is divided into five categories, each supporting different operations, and each piece of equipment has a different number of workstations. Cars provide general resources. Different operations require a car and corresponding equipment to be performed. We do not consider the car's transit time. The car and equipment can only be obtained by one operation at a time and released after completion. The experiment built two simulation scenarios for evaluation. 
\begin{figure}[tbh]
		\centering
		\includegraphics[width=8cm,height=4cm]{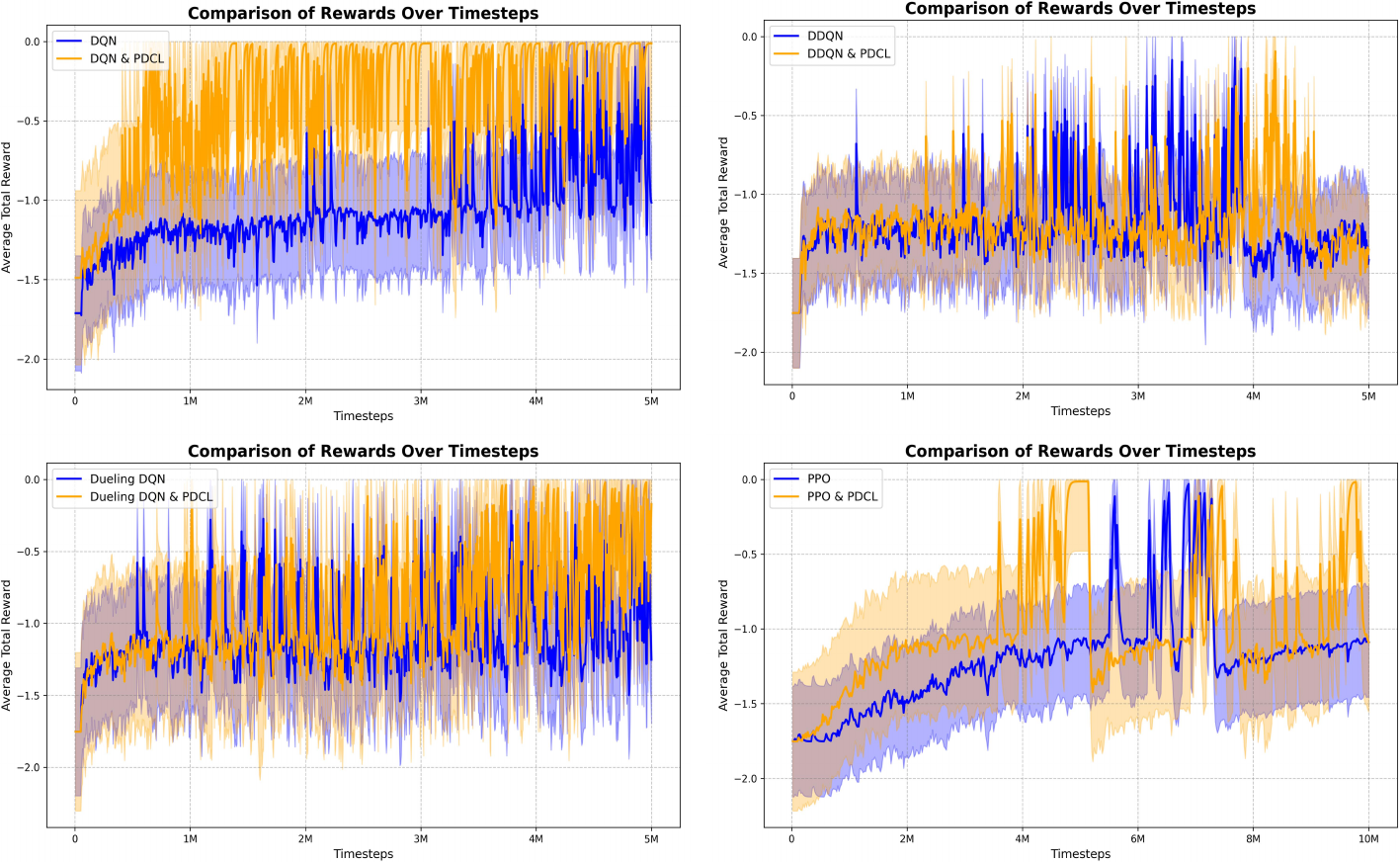}
		\caption{Results of different algorithms and joint pattern guidance on 10 tasks.}\label{3}
	\end{figure}
The first scenario includes ten tasks, three cars, and the number of workstations in the five pieces of equipment were two, two, two, one, and two. The second scenario included 12 tasks, and the number of workstations in the five pieces of equipment was the same as that in the first scenario. Machine learning methods are trained with Intel i9-14900K CPU, GTX4090 GPU a, 64GB RAM or Intel i7-13900K CPU, GTX4080 GPU a,32GB RAM. The same algorithm was tested on the same computer to ensure the readiness of the verification results.

\noindent\textbf{Evaluation metrics.} Completion time (ComT), cumulative reward (CumR), decision time (DecT), and training time (TraT) were used as metrics to measure the performance of the above methods, where ComT is the time slice length of tasks completed in one period in minutes. CumR is the total value of the rewards accumulated in one period. DecT is the total time required to assign a specified number of tasks using the trained model in milliseconds. TraT is the total time spent training for one period in minutes.

\noindent\textbf{Compared approaches.} The algorithms we use are implemented from \cite{stable-baselines3}. The lower-layer model training uniformly uses PPO, whereas the upper-layer model uses DQN, PPO, DDQN ,and Dueling DQN. We then added pattern mateching to the environment to guide retraining of the decision model. During training, all the models maintained fixed hyperparameters relative to themselves. PPO was trained for 10 million steps, with Tanh as the activation function, while DQN, DDQN, and Dueling DQN were trained for 5 million steps, with Sigmoid as the activation function. Except for Dueling DQN with a learning rate of 2e-5 in the 12-task scenario, the learning rate of other scenarios was uniformly set to 2e-4.

\subsection{Result Analysis}
\noindent\textbf{Simulation environment 1.} Table~\ref{tab:table1} lists the performance of the different algorithms and the addition of rpattern guidance in simulation environment 1. From the perspective of the cumulative completion time, after adding pattern guidance, the effects of DQN, DDQN, Dueling DQN, and PPO were improved by 9.4\%, 3.4\%, 1.8\%, and 6.4\%, respectively. Regarding training time, the time of each algorithm increased significantly compared with that without pattern guidance because the agent can conduct more exploration in each epoch under pattern guidance (Figure~\ref{3}). After adding pattern guidance during the training process, the average cumulative reward per round was much higher than that without adding pattern guidance. 
\begin{figure}[tbh]
		\centering
		\includegraphics[width=8cm,height=4cm]{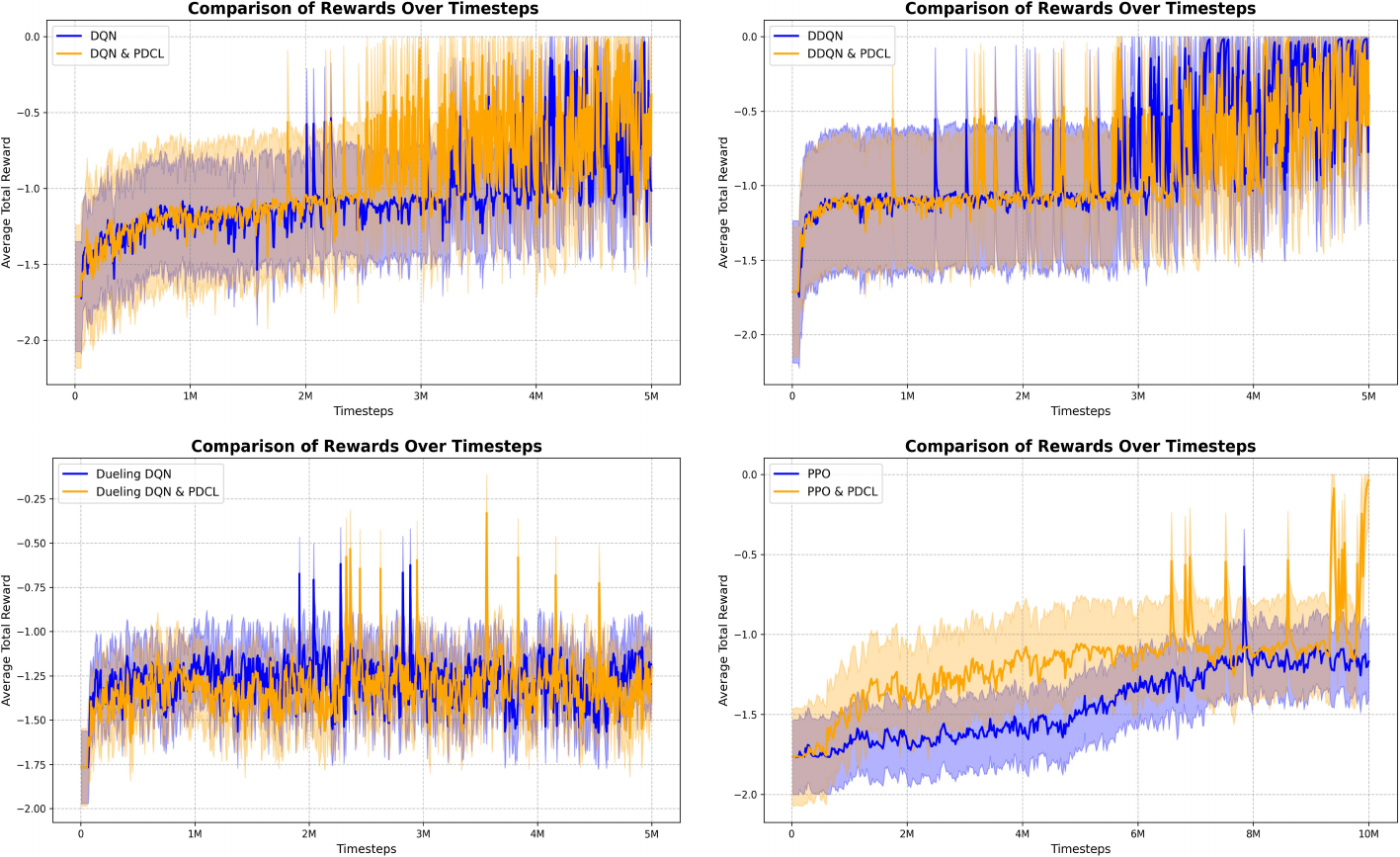}
		\caption{Results of different algorithms and joint pattern guidance on 12 tasks.}\label{4}
	\end{figure}
Simultaneously, regarding response time, after adding pattern guidance, it increased by 1.09 ms, 1.39 ms, 1.65 ms, and 1.1 ms, respectively, which did not affect the real-time performance of the algorithm. To improve performance, only a minimal compromise in real time was required.

\noindent\textbf{Simulation environment 2.} To further explore the effects of different operating conditions on CFL, we conducted a set of experiments by increasing the operating conditions from 10 to 12.
As shown in Table~\ref{tab:table1}, the completion times of DQN, DDQN, Dueling DQN, and PPO were reduced by 5 min, 2 min, 3.4 min, and 4 min, respectively. As shown in Figure~\ref{4}, after adding pattern guidance, the number of scheduling tasks completed by each algorithm was significantly improved compared with the number without pattern guidance.

\noindent\textbf{Additional experiment1.} We set the learning rate to 2e-5. As shown in Table~\ref{tab:table2}, after adding pattern guidance, the scheduling times of Dueling DQN and PPO reduced by 1 and 2 min, respectively, which proves that our framework remains effective after adjusting the learning rate. Although the effects of DQN and DDQN did not improve, we found that this was close to the optimal solution. However, the Appendi A shows that the speed and training stability significantly improved after adding pattern guidance.
	
\noindent\textbf{Additional experiment2.} We set the activation function to ReLU, and PPO was trained to change to 5 million steps. As shown in Table~\ref{tab:table3}, although the response and training times of each algorithm increased slightly after adding pattern guidance, the scheduling effect improved, which was acceptable. Notably, before adding pattern guidance, the Dueling DQN could not converge; however, after adding pattern guidance, it completed the scheduling task. Therefore, the model can better complete the scheduling task after learning the priority of the resource allocation. Simultaneously, after adjusting the hyperparameters, the pattern guidance remains effective.

\section{Conclusion}
We propose correctness learning incorporating deductive verification and historical high-quality schemes. Leveraging the reasoning capabilities of separation logic applied to shared resources, we model and reason the task priorities hidden in the scheme. The matching degree between the task priorities in the model output scheme and the task priorities in the high-quality scheme guides the training process, which effectively combines the historical experience with the exploratory capability of the reinforcement learning.
The experimental results demonstrate that our method enables multiple algorithms to surpass their original performance in diverse conditions. In the future we will identify more general pattern by analyzing more historical high-quality schemes. Additionally, we will investigate the different importance of different pattern within the overall decision-making framework.

\bibliographystyle{named}
\bibliography{ijcai25}

\begin{thebibliography}{}

\bibitem[\protect\citeauthoryear{Albarghouthi \bgroup \em et al.\egroup }{2021}]{albarghouthi2021introduction}
Aws Albarghouthi, Zidong Zhang, and Dongxia Zhang.
\newblock Introduction to neural network verification.
\newblock {\em Foundations and Trends{\textregistered} in Programming Languages}, 7(1--2):1--157, 2021.

\bibitem[\protect\citeauthoryear{Bai \bgroup \em et al.\egroup }{2023}]{bai2023achieving}
Qinbo Bai, Amrit~Singh Bedi, and Vaneet Aggarwal.
\newblock Achieving zero constraint violation for constrained reinforcement learning via conservative natural policy gradient primal-dual algorithm.
\newblock In {\em Proceedings of the AAAI Conference on Artificial Intelligence}, volume~37, pages 6737--6744, 2023.

\bibitem[\protect\citeauthoryear{Beard and Baheri}{2022}]{beard2022black}
Jared~J Beard and Ali Baheri.
\newblock Black-box safety validation of autonomous systems: A multi-fidelity reinforcement learning approach.
\newblock {\em arXiv preprint arXiv:2203.03451}, 2022.

\bibitem[\protect\citeauthoryear{Bharadhwaj \bgroup \em et al.\egroup }{2020}]{bharadhwaj2020conservative}
Homanga Bharadhwaj, Aviral Kumar, Nicholas Rhinehart, Sergey Levine, Florian Shkurti, and Animesh Garg.
\newblock Conservative safety critics for exploration.
\newblock {\em arXiv preprint arXiv:2010.14497}, 2020.

\bibitem[\protect\citeauthoryear{Brooks \bgroup \em et al.\egroup }{2023}]{brooks2023large}
Ethan Brooks, Logan Walls, Richard~L Lewis, and Satinder Singh.
\newblock Large language models can implement policy iteration.
\newblock {\em Advances in Neural Information Processing Systems}, 36:30349--30366, 2023.

\bibitem[\protect\citeauthoryear{Cao \bgroup \em et al.\egroup }{2024}]{cao2024enhancing}
Chengzhi Cao, Yinghao Fu, Sheng Xu, Ruimao Zhang, and Shuang Li.
\newblock Enhancing human-ai collaboration through logic-guided reasoning.
\newblock In {\em The Twelfth International Conference on Learning Representations}, 2024.

\bibitem[\protect\citeauthoryear{Carr \bgroup \em et al.\egroup }{2023}]{carr2023safe}
Steven Carr, Nils Jansen, Sebastian Junges, and Ufuk Topcu.
\newblock Safe reinforcement learning via shielding under partial observability.
\newblock In {\em Proceedings of the AAAI Conference on Artificial Intelligence}, volume~37, pages 14748--14756, 2023.

\bibitem[\protect\citeauthoryear{Chakraborty \bgroup \em et al.\egroup }{2024}]{chakraborty2024parl}
Souradip Chakraborty, Amrit Bedi, Alec Koppel, Huazheng Wang, Dinesh Manocha, Mengdi Wang, and Furong Huang.
\newblock Parl: A unified framework for policy alignment in reinforcement learning from human feedback.
\newblock In {\em The Twelfth International Conference on Learning Representations}, 2024.

\bibitem[\protect\citeauthoryear{Christiano \bgroup \em et al.\egroup }{2017}]{christiano2017deep}
Paul~F Christiano, Jan Leike, Tom Brown, Miljan Martic, Shane Legg, and Dario Amodei.
\newblock Deep reinforcement learning from human preferences.
\newblock {\em Advances in neural information processing systems}, 30, 2017.

\bibitem[\protect\citeauthoryear{Crochepierre \bgroup \em et al.\egroup }{2022}]{crochepierre2022interactive}
Laure Crochepierre, Lydia Boudjeloud-Assala, and Vincent Barbesant.
\newblock Interactive reinforcement learning for symbolic regression from multi-format human-preference feedbacks.
\newblock In {\em IJCAI}, pages 5900--5903, 2022.

\bibitem[\protect\citeauthoryear{Fu \bgroup \em et al.\egroup }{2018}]{fu2018model}
Chen Fu, Andrea Turrini, Xiaowei Huang, Lei Song, Yuan Feng, and Lijun Zhang.
\newblock Model checking probabilistic epistemic logic for probabilistic multiagent systems.
\newblock In {\em IJCAI International Joint Conference on Artificial Intelligence}, 2018.

\bibitem[\protect\citeauthoryear{Ghosh}{2023}]{ghosh2023interpretability}
Bishwamittra Ghosh.
\newblock Interpretability and fairness in machine learning: A formal methods approach.
\newblock In {\em IJCAI}, pages 7083--7084, 2023.

\bibitem[\protect\citeauthoryear{Hunt \bgroup \em et al.\egroup }{2021}]{hunt2021verifiably}
Nathan Hunt, Nathan Fulton, Sara Magliacane, Trong~Nghia Hoang, Subhro Das, and Armando Solar-Lezama.
\newblock Verifiably safe exploration for end-to-end reinforcement learning.
\newblock In {\em Proceedings of the 24th International Conference on Hybrid Systems: Computation and Control}, pages 1--11, 2021.

\bibitem[\protect\citeauthoryear{Kaufmann \bgroup \em et al.\egroup }{2023}]{kaufmann2023survey}
Timo Kaufmann, Paul Weng, Viktor Bengs, and Eyke H{\"u}llermeier.
\newblock A survey of reinforcement learning from human feedback.
\newblock {\em arXiv preprint arXiv:2312.14925}, 2023.

\bibitem[\protect\citeauthoryear{Kouvaros}{2023}]{kouvaros2023towards}
Panagiotis Kouvaros.
\newblock Towards formal verification of neuro-symbolic multi-agent systems.
\newblock In {\em IJCAI}, pages 7014--7019, 2023.

\bibitem[\protect\citeauthoryear{Krichen \bgroup \em et al.\egroup }{2022}]{krichen2022formal}
Moez Krichen, Alaeddine Mihoub, Mohammed~Y Alzahrani, Wilfried Yves~Hamilton Adoni, and Tarik Nahhal.
\newblock Are formal methods applicable to machine learning and artificial intelligence?
\newblock In {\em 2022 2nd International Conference of Smart Systems and Emerging Technologies (SMARTTECH)}, pages 48--53. IEEE, 2022.

\bibitem[\protect\citeauthoryear{Landers and Doryab}{2023}]{landers2023deep}
Matthew Landers and Afsaneh Doryab.
\newblock Deep reinforcement learning verification: a survey.
\newblock {\em ACM Computing Surveys}, 55(14s):1--31, 2023.

\bibitem[\protect\citeauthoryear{Li \bgroup \em et al.\egroup }{2022}]{li2022review}
Chunquan Li, Yaqiong Chen, and Yuling Shang.
\newblock A review of industrial big data for decision making in intelligent manufacturing.
\newblock {\em Engineering Science and Technology, an International Journal}, 29:101021, 2022.

\bibitem[\protect\citeauthoryear{Mansouri \bgroup \em et al.\egroup }{2023}]{mansouri2023iot}
Seyed~Amir Mansouri, Ahmad~Rezaee Jordehi, Mousa Marzband, Marcos Tostado-V{\'e}liz, Francisco Jurado, and Jos{\'e}~A Aguado.
\newblock An iot-enabled hierarchical decentralized framework for multi-energy microgrids market management in the presence of smart prosumers using a deep learning-based forecaster.
\newblock {\em Applied Energy}, 333:120560, 2023.

\bibitem[\protect\citeauthoryear{Odriozola-Olalde \bgroup \em et al.\egroup }{2023}]{odriozola2023shielded}
Haritz Odriozola-Olalde, Maider Zamalloa, and Nestor Arana-Arexolaleiba.
\newblock Shielded reinforcement learning: A review of reactive methods for safe learning.
\newblock In {\em 2023 IEEE/SICE International Symposium on System Integration (SII)}, pages 1--8. IEEE, 2023.

\bibitem[\protect\citeauthoryear{Raffin \bgroup \em et al.\egroup }{2021}]{stable-baselines3}
Antonin Raffin, Ashley Hill, Adam Gleave, Anssi Kanervisto, Maximilian Ernestus, and Noah Dormann.
\newblock Stable-baselines3: Reliable reinforcement learning implementations.
\newblock {\em Journal of Machine Learning Research}, 22(268):1--8, 2021.

\bibitem[\protect\citeauthoryear{Sokol and Vogt}{2024}]{sokol2024does}
Kacper Sokol and Julia~E Vogt.
\newblock What does evaluation of explainable artificial intelligence actually tell us? a case for compositional and contextual validation of xai building blocks.
\newblock In {\em Extended Abstracts of the CHI Conference on Human Factors in Computing Systems}, pages 1--8, 2024.

\bibitem[\protect\citeauthoryear{Verma \bgroup \em et al.\egroup }{2019}]{verma2019imitation}
Abhinav Verma, Hoang Le, Yisong Yue, and Swarat Chaudhuri.
\newblock Imitation-projected programmatic reinforcement learning.
\newblock {\em Advances in Neural Information Processing Systems}, 32, 2019.

\bibitem[\protect\citeauthoryear{Visan \bgroup \em et al.\egroup }{2022}]{visan2022towards}
Maria Visan, Sorin~Lenus Negrea, and Firicel Mone.
\newblock Towards intelligent public transport systems in smart cities; collaborative decisions to be made.
\newblock {\em Procedia Computer Science}, 199:1221--1228, 2022.

\bibitem[\protect\citeauthoryear{Wolfman and Weld}{1999}]{wolfman1999lpsat}
Steven~A Wolfman and Daniel~S Weld.
\newblock The lpsat engine \& its application to resource planning.
\newblock In {\em IJCAI}, volume 1999, pages 310--317. Citeseer, 1999.

\bibitem[\protect\citeauthoryear{Xiao \bgroup \em et al.\egroup }{2024}]{xiao2024policy}
Minheng Xiao, Xian Yu, and Lei Ying.
\newblock Policy gradient methods for risk-sensitive distributional reinforcement learning with provable convergence.
\newblock {\em arXiv preprint arXiv:2405.14749}, 2024.

\bibitem[\protect\citeauthoryear{Yujie \bgroup \em et al.\egroup }{2023}]{yujie2023optimization}
LIU Yujie, HAN Wei, SU~Xichao, and CUI Rongwei.
\newblock Optimization of fixed aviation support resource station configuration for aircraft carrier based on aircraft dispatch mission scheduling.
\newblock {\em Chinese Journal of Aeronautics}, 36(2):127--138, 2023.

\bibitem[\protect\citeauthoryear{Zhang \bgroup \em et al.\egroup }{2024}]{zhang2024multi}
Natalia Zhang, Xinqi Wang, Qiwen Cui, Runlong Zhou, Sham~M Kakade, and Simon~S Du.
\newblock Multi-agent reinforcement learning from human feedback: Data coverage and algorithmic techniques.
\newblock {\em arXiv preprint arXiv:2409.00717}, 2024.

\end{thebibliography}

\end{document}